\documentclass{article}

\PassOptionsToPackage{numbers, compress}{natbib}
\usepackage[final]{nips_2017}
\usepackage{algpseudocode}
\usepackage{algorithm}
\usepackage{amsmath}
\usepackage{amssymb}
\usepackage{graphicx}
\DeclareGraphicsExtensions{.pdf,.png,.jpg}
\usepackage{subcaption}
\usepackage{multirow}
\usepackage{bm}

\usepackage[utf8]{inputenc} 
\usepackage[T1]{fontenc}    
\usepackage{hyperref}       
\usepackage{url}            
\usepackage{booktabs}       
\usepackage{amsfonts}       
\usepackage{nicefrac}       
\usepackage{microtype}      

\title{Generative Adversarial Trainer: Defense to Adversarial Perturbations with GAN}

\author{
  Hyeungill Lee, \quad Sungyeob Han, \quad Jungwoo Lee \\
  Department of electrical and computer engineering \\
  Seoul National University \\
  \texttt{\{hilee, syhan\}@wspl.snu.ac.kr}, \texttt{ junglee@snu.ac.kr} \\
}

\begin{document}
\maketitle
\begin{abstract}
We propose a novel technique to make neural network robust to adversarial examples using a generative adversarial network. We alternately train both classifier and generator networks. The generator network generates an adversarial perturbation that can easily fool the classifier network by using a gradient of each image. Simultaneously, the classifier network is trained to classify correctly both original and adversarial images generated by the generator. These procedures help the classifier network to become more robust to adversarial perturbations. Furthermore, our adversarial training framework efficiently reduces overfitting and outperforms other regularization methods such as Dropout. We applied our method to supervised learning for CIFAR datasets, and experimental results show that our method significantly lowers the generalization error of the network. To the best of our knowledge, this is the first method which uses GAN to improve supervised learning.
\end{abstract}

\section{Introduction}
Recently, deep learning has advanced in all areas of artificial intelligence, including image classification and speech recognition \citep{hinton2012deep, krizhevsky2012imagenet}. These advances were owing to deep neural networks which can be easily trained by backpropagation, so they can represent complex probability distributions over high dimensional data. Despite these advances, deep neural networks remain imperfect. In particular, they show weaknesses with respect to adversarial examples when compared to humans \citep{szegedy2013intriguing}. Adversarial examples can effectively fool a neural network to change its predictions, and the human eye cannot distinguish such examples from original images.

Several studies have attempted to make neural networks robust to such adversarial examples \citep{goodfellow2014explaining, miyato2015distributional, papernot2016distillation}. Adversarial training is one of the methods that retrains a neural network to predict correct labels for adversarial examples. In adversarial training, adversarial examples are generated in the inner loop of the adversarial training algorithms. Thus, this process should be fast to help adversarial training to be practical. \citet{goodfellow2014explaining} proposed the fast gradient sign method, which is a simple and fast method of generating adversarial examples. 

\citet{goodfellow2014generative} also introduced generative adversarial networks (GAN). GAN is a framework to train generative models, and shows a state-of-the-art performance for image generation \citep{berthelot2017began, radford2015unsupervised}. The main idea of GAN is that two networks play a minimax game so that they converge gradually to an optimal solution.

In this paper, we propose a novel adversarial training method using a GAN framework. Similar to GAN, we alternately train both a classifier network (trainee) and a generator network (trainer). The generator network attempts to generate adversarial perturbations that can easily fool the classifier network, whereas the classifier network tries to classify correctly both original and adversarial images produced by the generator network. These procedures gradually help the classifier network to be robust to adversarial perturbations. Our experimental results show that our method outperforms other adversarial training using a fast gradient method. We also observe that generalization errors are remarkably lower than that of other regularization methods such as dropout \citep{srivastava2014dropout}.

\section{Backgrounds}
In this section, we briefly review the adversarial training (with fast gradient method) and GAN. 

\subsection{Adversarial Training}
\citet{goodfellow2014explaining} introduced a rational explanation for the adversarial example, and proposed a fast technique to generate adversarial perturbation. The authors observed that adversarial examples exist because models are too linear. They suggested a fast gradient sign method such that

\begin{equation}
    \eta = \epsilon \text{sign}(\nabla_{\bm{x}}J(\bm{\theta},\bm{x},y))
\end{equation}

where $\eta$ is an adversarial perturbation, $\bm{\theta}$ denotes parameters of the network, $\bm{x}$ is the input with the label $y$, and $J(\bm{\theta},\bm{x},y)$ denotes the cost function to train the classifier network.

Deep neural networks trained by standard supervised methods are vulnerable to adversarial examples. Adversarial training helps neural networks to be robust to adversarial perturbation. The practical method of adversarial training is to introduce an adversarial objective function using fast gradient sign method. Let $\tilde{J}(\bm{\theta},\bm{x},y)$ be the loss function of adversarial training, which optimizes a neural network against an adversary.

\begin{equation} 
\label{goodfellow}
  \tilde{J}(\bm{\theta},\bm{x},y)=\alpha J(\bm{\theta},\bm{x},y)+(1-\alpha)J(\bm{\theta},\bm{x}+\epsilon \text{sign}(\nabla_{\bm{x}} J(\bm{\theta},\bm{x},y)),y)
\end{equation}

Eq. (\ref{goodfellow}) consists of two cost functions. The first cost function is the original cross-entropy loss function for a neural network. The second is the loss function with adversarial perturbations, which is added to each input $\bm{x}$. Note that $\alpha$ is a hyperparameter that adjusts the ratio between the two cost functions. In \citep{shaham2015understanding}, the authors analyzed the principle of adversarial training and found a strong connection between robust optimization and regularization. This establishes a minimization-maximization approach for adversarial training, which in turn makes neural networks stable in a neighborhood around training points. The authors also generalized the fast gradient sign method and proposed another adversarial training method with $L_2$ norm constraint \citep{shaham2015understanding}. 

\begin{equation}
\label{fgl2}
    \eta = \epsilon \frac{\nabla_{\bm{x}} J(\bm{\theta},\bm{x},y)}{\| \nabla_{\bm{x}} J(\bm{\theta},\bm{x},y)\|_2}
\end{equation}

\begin{equation}
\label{l2_method}
  \tilde{J}(\bm{\theta},\bm{x},y)=\alpha J(\bm{\theta},\bm{x},y)+(1-\alpha)J(\bm{\theta},\bm{x}+\epsilon \frac{\nabla_{\bm{x}} J(\bm{\theta},\bm{x},y)}{\| \nabla_{\bm{x}} J(\bm{\theta},\bm{x},y)\|_2},y)
\end{equation}

\subsection{Generative Adversarial Networks}
GAN \citep{goodfellow2014generative} is a new successful framework for generative models. The conventional means of training a generative model is to maximize the likelihood function, which computes various quantities such as marginal probabilities and partition functions, which are computationally intractable in most cases. GAN allows us to train a generative model without the intractable computation. A GAN framework forces two networks to compete with each other. These two networks are: a generative model which attempts to map a sample $z$ (noise distribution) to the data distribution, and a discriminative model, which attempts to discriminate between training data and a sample from a generative model. The goal of a generative model is to maximize the probability that the discriminative model will produce a mistake. Thus, generative and discriminative models play the following two-player minimax game with a value function $V(G,D)$

\begin{equation}
\label{gan_eq}
    \underset{G}{\text{min}} \underset{D}{\text{max}} V(D,G)=E_{x \sim p_{data}(x)}[\log D(x)]+E_{z \sim p_z(z)}[\log (1-D(G(z)))]
\end{equation}
The competition in this minimax game forces both models to improve their ability until the discriminator cannot distinguish a generated sample from a data sample.

\begin{figure}[h]
\centering
\includegraphics[width=1.0\linewidth]{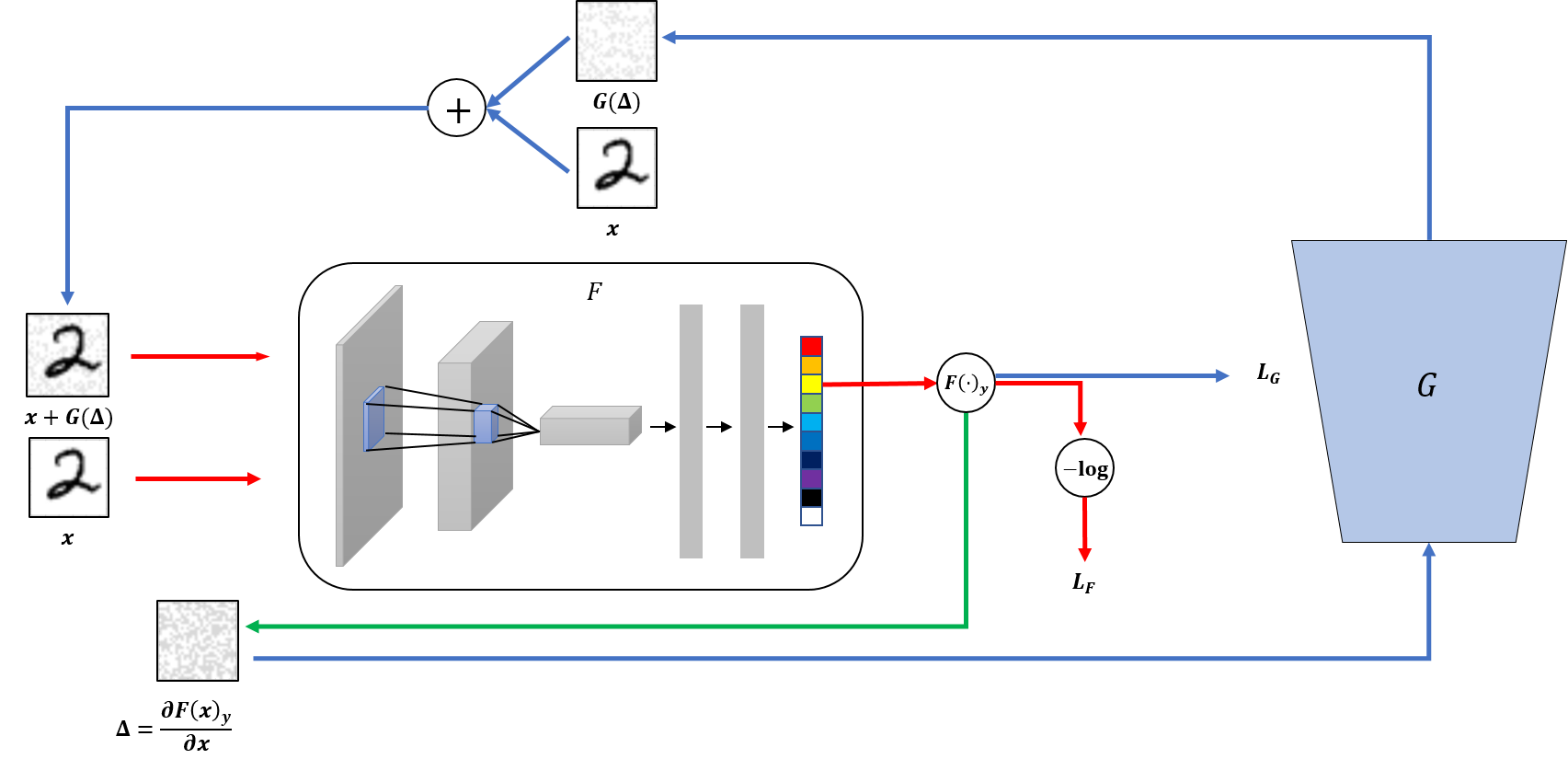}
\caption{\textbf{Adversarial training with Generative Adversarial Trainer:} (1) Generative Adversarial Trainer $G$ is trained to generate an adversarial perturbation that can fool the classifier network using the gradient of each image. (2) Classifier Network $F$ is trained to classify correctly both original and adversarial examples generated by $G$.}
\label{fig1}
\end{figure}

\section{Proposed Method}
\label{headings}
In this section, we propose a novel adversarial training framework. However, we first introduce a generative adversarial trainer (GAT), which plays a major role in adversarial training. The objective of the GAT is to generate adversarial perturbations that can easily fool the classifier network using a gradient of images. A classifier is trained to classify correctly both original and adversarial images generated by the GAT. The entire procedure is shown in Fig. \ref{fig1}. In Section \ref{notation}, we describe our notation. Section \ref{gatsec} describes the structure of the GAT, Section \ref{atgat} explains the adversarial training mechanism used with the GAT.

\subsection{Notations}
\label{notation}
We denote a labeled training set by ${\{(\bm{x}^{(i)},y^{(i)})\}}^N_{i=1}$, where $\bm{x}^{(i)} \in \mathbb{R}^{H\times W \times C}$ represents input images with height $H$, width $W$, and channel $C$, and $y^{(i)} \in \{1, \cdots, K\}$ is a label for an input $\bm{x}^{(i)}$.
We use two neural networks in the proposed method. One is a standard K-class classifier network $F(\bm{x};\theta_f)$ which is defined by:
\begin{equation}
\label{fx}
  F : \mathbb{R}^{H \times W \times C} \rightarrow \mathbb{R}^K, \quad F(\bm{x}) = [F(\bm{x})_1, F(\bm{x})_2, \cdots, F(\bm{x})_K]^T 
\end{equation}
where $F(\bm{x})$ represents the class probability vector computed using the softmax function. The other is a GAT $G(\Delta;\theta_g)$ which is defined by:
\begin{equation}
\label{gx}
  G : \mathbb{R}^{H \times W \times C} \rightarrow \mathbb{R}^{H \times W \times C}
\end{equation}
Note that $G(\Delta)$ represents the perturbation of the input image $\bm{x}$, where $\Delta = {\frac{\partial F(\bm{x})_y}{\partial \bm{x}}}$ denotes the gradient of input images with respect to the class probability of the label. We use a cross entropy loss function for the classifier $F(\bm{x};\theta_f)$, which is denoted by:
\begin{equation}
J(\theta_f,\bm{x},y)= - \log{F(\bm{x};\theta_f)_{y}}
\end{equation}

\subsection{Generative Adversarial Trainer}
\label{gatsec}
The main idea of the GAT is to use a neural network to find the perturbation generator function specific to the classifier rather than just the sign or normalized functions used in the fast gradient method. The objective of the GAT is to find best perturbation image using the gradient of each image. To achieve this goal, the loss function is defined as follows:
\begin{equation}
\label{lg}
L_G(\Delta, y) = F(\bm{x}+G(\Delta))_y + c_g \cdot \big\|G(\Delta)\big\|_2^2
\end{equation}
The loss function of GAT consists of the two cost functions. One is the loss function, which is used to find perturbation images that lower the classifier's class probability. The other cost function restricts the power of the perturbation so that it is not too large. In (\ref{lg}), $c_g$ is a hyperparameter that adjusts the ratio between two cost functions. If $c_g$ is too low, it will find only a trivial solution with very high perturbation power. If $c_g$ is too high, it will generate only a zero perturbation image. Therefore, finding the appropriate $c_g$ through a hyperparameter search is crucial.

\subsection{Adversarial Training with GAT}
\label{atgat}
As an analogy, our adversarial training framework is similar to the spring training of a baseball team. A trainer analyzes the vulnerable points of a player and, based on this analysis, trains his weakest parts in addition to providing general training. This process is repeated over and over again. The goal at the end of the spring camp, is that the player overcomes most of the weaknesses and becomes a better player.

GAT plays a similar role to a trainer for a baseball team. During each training step, GAT learns to generate the best adversarial perturbation for each input. Simultaneously, a classifier network is trained to classify correctly both original and adversarial examples generated by GAT. The loss function of GAT is given as (\ref{lg}), whereas that of the classifier network is based on the adversarial objective function.

\begin{equation}
\label{lf}
  L_F = \alpha \cdot J(\theta_f,\bm{x},y)+(1-\alpha) \cdot J(\theta_f, \bm{x}+G(\Delta),y)
\end{equation}
 
For the simplicity's sake, we used $\alpha=0.5$ in all experiments. Similar to GAN, completely optimizing GAT in the inner loop of training is computationally expensive and would result in overfitting if we do not have a large number of datasets. Instead, we alternately optimize generator network $k$ steps and the classifier network $1$ step. GAT is maintained near its optimal solution if the classifier network varies in a sufficiently slow enough. We used $k=1$ in all of our experiments. The entire procedure is presented in Algorithm \ref{alg:algorithm1}.

\begin{algorithm}
\caption{Adversarial Training with GAT}
\label{alg:algorithm1}
\textbf{Input:} training data $\{(\bm{x}^{(i)},y^{(i)})\}_{i=1}^{N}$, classifier network $F(x;\theta_f)$, generator network $G(x;\theta_g)$\\
\bigskip
\textbf{Output:} Robust parameter vector $\theta_f$\\
initialize $\theta_f, \theta_g$
\begin{algorithmic}
\For{number of training iterations}
\For{k steps}
\State Sample minibatch of $m$ examples $\{(\bm{x}^{(1)},y^{(1)}),...,(\bm{x}^{(m)},y^{(m)})\}$ from data distribution.
\State Update the generator by descending its stochastic gradient:
\begin{equation} \notag
  \nabla_{\theta_g} {\frac{1}{m}} \sum_{i=1}^{m} \biggl[F\big(x^{(i)}+G(\Delta^{(i)})\big)_{y^{(i)}}+ c_g \cdot \big\|G(\Delta^{(i)})\big\|_2^2\biggl]
\end{equation}

\EndFor
\\
\State Sample minibatch of $m$ examples  $\{(\bm{x}^{(1)},y^{(1)}),...,(\bm{x}^{(m)},y^{(m)})\}$ from data distribution.
\State Update the classifier by descending its stochastic gradient:
\begin{equation} \notag
  \nabla_{\theta_f} {\frac{1}{m}} \sum_{i=1}^{m} \biggl[\alpha \cdot J\big(\theta_f, \bm{x}^{(i)}, y^{(i)}\big)+(1-\alpha) \cdot J\big(\theta_f, \bm{x}^{(i)}+G(\Delta^{(i)}), y^{(i)}\big)\biggl]
\end{equation}

\EndFor
\\
\State The gradient-based updates can use any standard gradient-based learning rule. We used the Adam-Optimizer in our experiments.
\end{algorithmic}
\end{algorithm}

\section{Experiments}
To verify the performance of our proposed method, we experimented on two CIFAR datasets \citep{krizhevsky2009learning}: CIFAR-10 and CIFAR-100, which consist of $60,000$ $32 \times 32$ colour images in 10 or 100 classes, respectively, with $6,000$ images per class. The two datasets each contain $50,000$ training samples and $10,000$ test samples. We split the original $50,000$ training samples into $45,000$ training samples and $5,000$ validation samples, and used the latter to tune the hyperparameters. We performed all experiments using TensorFlow \citep{abadi2015tensorflow}.

\subsection{Experimental setup}

The network architectures used in our experiments are described in Table \ref{tab}. We used a small version of Allconvnet proposed by \citet{springenberg2014striving} as a classifier (trainee) because this model has a simple structure and yields good performance. The input image to our ConvNets was a fixed-size $32 \times 32 \times 3$ and normalized between $0$ and $1$. We used $3 \times 3$ convolutions with rectified linear units (ReLUs) as activation functions. The number of convolutional filters were increased linearly with each down-sampling which was implemented as sub-sampling with stride 2. Fully connected layers were replaced by simple $1 \times 1$ convolutions and the scores of each class were averaged over the spatial dimensions.

The generator network (trainer) had a simpler structure. The input gradient image of the generator network was computed from the classifier network. The network consisted of six $3 \times 3$ convolutional layers followed by two $1 \times 1$ convolutional layers, and used hyperbolic tangent as an activation function of the output layer. 

Our model had no batch normalization \citep{ioffe2015batch}, no dropout \citep{srivastava2014dropout}, and no weight decay. These techniques may have slightly improved our results but were not a major concern of our study. We trained our model using the Adam optimizer \citep{kingma2014adam} with a learning rate $1 \times 10^{-3}$ for the classifier and $1 \times 10^{-6}$ for the generator. The network weights were initialized using Xavier initialization method \citep{glorot2010understanding}, which improved the speed of convergence.

\begin{table}[h]
\caption{Model description for CIFAR datasets}
\label{tab}
\begin{center}
\begin{tabular}{c c}
\hline
classifier network (trainee) & generator network (trainer) \\
\hline
Input: $32 \times 32 \times 3$ RGB image & Input: $32 \times 32 \times 3$ gradient image \\
\hline
$3 \times 3$ conv $48$ ReLu & $3 \times 3$ conv $48$ ReLu \\
$3 \times 3$ conv $48$ ReLu, stride=$2$ & $3 \times 3$ conv $48$ ReLu \\ 
$3 \times 3$ conv $96$ ReLu & $3 \times 3$ conv $48$ ReLu \\ 
$3 \times 3$ conv $96$ ReLu, stride=$2$ & $3 \times 3$ conv $48$ ReLu \\ 
$3 \times 3$ conv $96$ ReLu & $3 \times 3$ conv $48$ ReLu \\ 
$1 \times 1$ conv $96$ ReLu & $3 \times 3$ conv $48$ ReLu \\ 
$1 \times 1$ conv $10$ (or $100$) & $1 \times 1$ conv $48$ ReLu \\
global averaging over $8 \times 8$ image &  $1 \times 1$ conv $3$ \\
softmax & tanh\\
\hline
Output: $10$ (or $100$) class probabilities & Output: $32 \times 32 \times 3$ perturbation image \\
\hline
\end{tabular}
\end{center}
\end{table}

\subsection{Perturbations generated by GAT}
\begin{figure}[h]
\centering
\begin{subfigure}{.5\textwidth}
  \centering
  \includegraphics[width=1.0\linewidth]{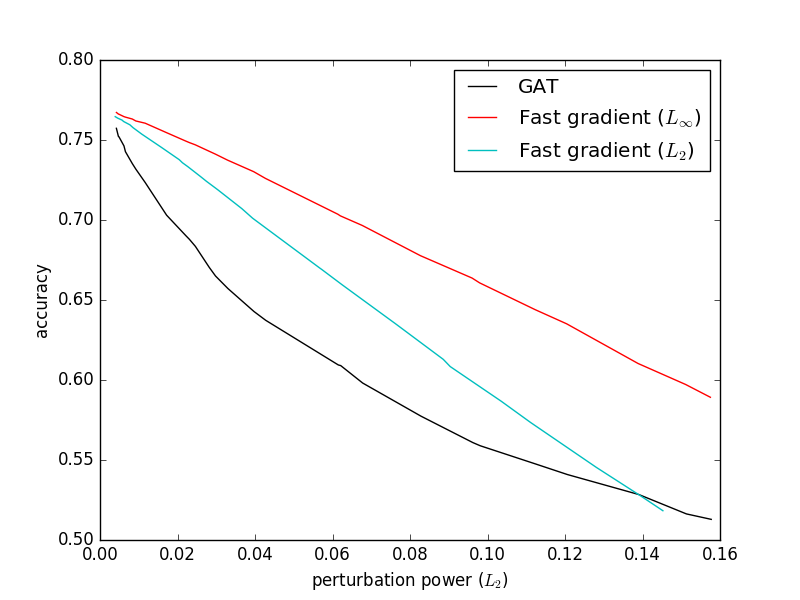}
  \caption{CIFAR-10}
\end{subfigure}%
\begin{subfigure}{.5\textwidth}
  \centering
  \includegraphics[width=1.0\linewidth]{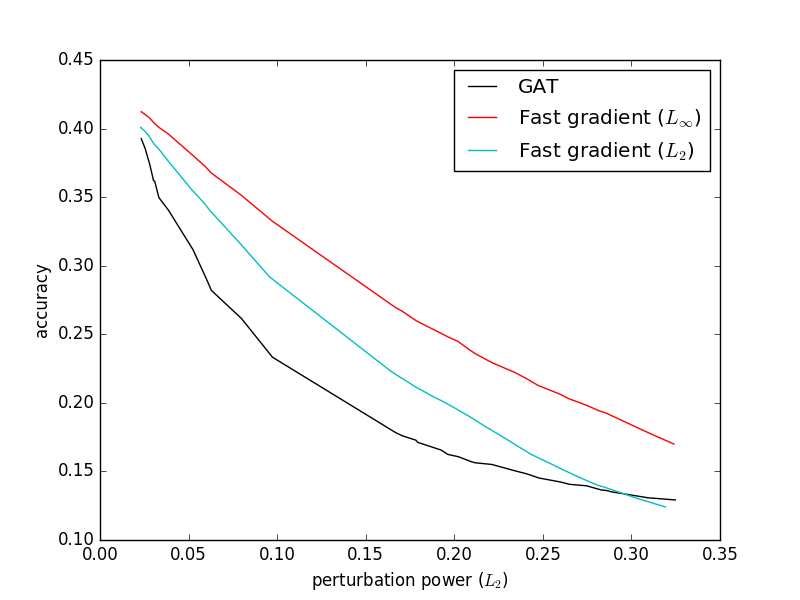}
  \caption{CIFAR-100}
\end{subfigure}
\caption{Comparison of the fast gradient attack (with $L_2$ and $L_\infty$ norm constraint) and our proposed GAT. Adversarial examples were generated for use with each method with various perturbation powers, and classification accuracy is plotted as shown.}
\label{fig2}
\end{figure}

Is it possible to find stronger perturbation images than that generated by fast gradient method when we only know the gradient of the original images? Surprisingly, the answer is yes. To verify this, we performed the following experiment.
First, we trained a ConvNet without using an adversarial training algorithm. We call this network a "baseline network". The classification accuracy of the baseline network was approximately 77\% and 44\% for CIFAR-10 and CIFAR-100 datasets, respectively.
We next trained our generator network (GAT) using the loss function given in (\ref{lg}). We used early stopping with validation loss, and the accuracy of the adversarial images generated by the generator network was computed using the test images.
We compared the accuracy of the adversarial images generated by the fast gradient method ($L_2$ or $L_\infty$) with the same perturbation power generated by the generator network.
Experiments were repeated with different $c_g$ values given in (\ref{lg}), so that adversarial perturbations with various powers could be generated. The results are shown in Fig. \ref{fig2}.

The generator network efficiently generates stronger adversarial images than those generated by the other methods when the perturbation power is low. Unlike the fast gradient method, the generator network requires several iterations to achieve optimal state. However, in adversarial training phase, full optimization of the generator in each training step is not required because we alternately optimize the generator and classifier network.

\subsection{Adversarial training with GAT}
Mainly two types of attacks against deep neural networks exist. One is a direct attack which generates adversarial examples based on a precise understanding of the model internals or its training data. The other is an indirect attack, which generates adversarial examples without knowing precise information about the model. We experimented with adversarial examples generated in these two cases to assess the robustness of our network against various attacks.

\subsubsection{Direct attack}

\begin{figure}[b]
\centering
\begin{subfigure}{.5\textwidth}
  \centering
  \includegraphics[width=1.0\linewidth]{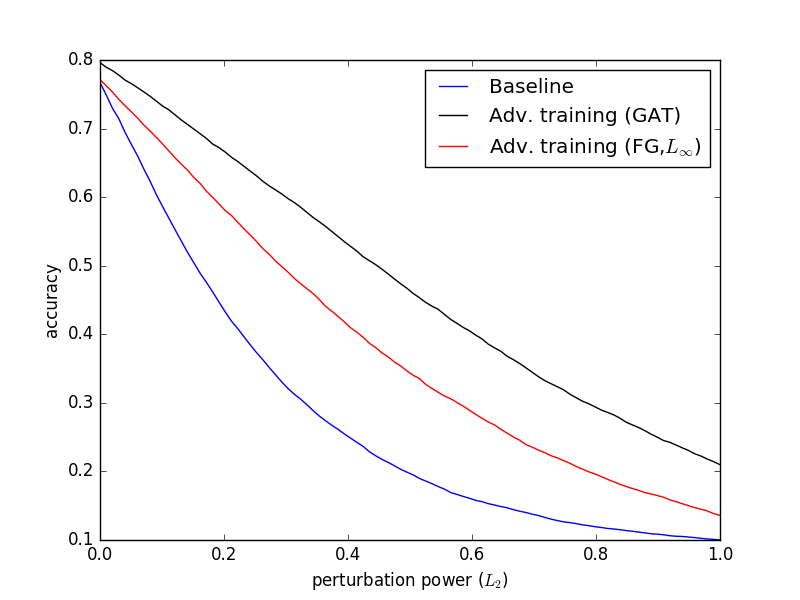}
  \caption{CIFAR10}
\end{subfigure}%
\begin{subfigure}{.5\textwidth}
  \centering
  \includegraphics[width=1.0\linewidth]{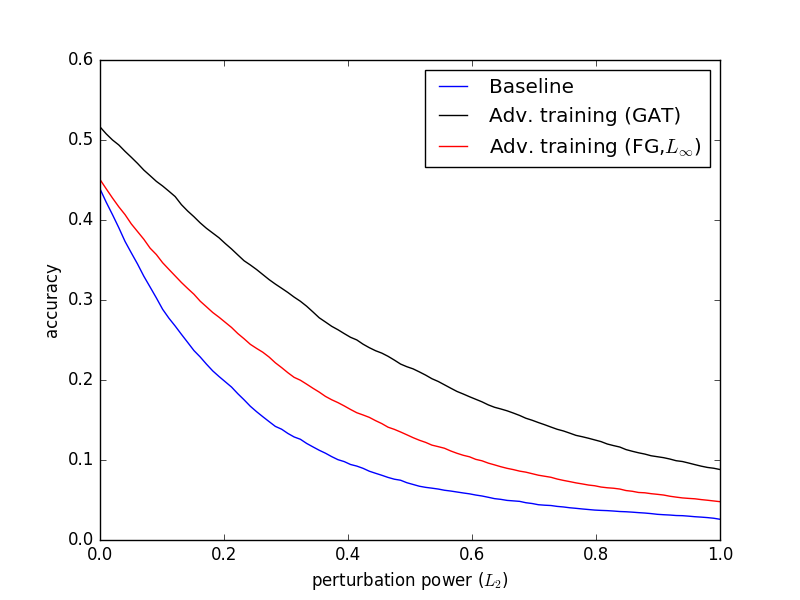}
  \caption{CIFAR100}
\end{subfigure}
\caption{Comparison of the baseline and robustified networks. Adversarial examples were generated using (\ref{fgl2}) from each network for various values of $\epsilon$, and classification accuracy is plotted as shown.}
\label{fig3}
\end{figure}

We compared the performance of the proposed adversarial training with the baseline network and the network trained with an adversarial objective function based on the fast gradient sign method. For each network, adversarial examples with various perturbation powers were generated by the fast gradient method with $L_2$ norm constraint using (\ref{fgl2}). Note that adversarial examples were generated for each network. Over the various perturbation powers, we measured the classification accuracy of each network for adversarial examples that were generated from each network's parameters. The results are shown in Fig. \ref{fig3}.
The proposed method (adversarial training with GAT) showed the best classification accuracy without perturbation. These results showed that our algorithm is also effective in regularizing the neural network. Furthermore, they showed that our method is more robust against adversarial perturbations than is the fast gradient sign method.

\subsubsection{Indirect attack}
\begin{figure}[h]
\centering
\begin{subfigure}{.5\textwidth}
  \centering
  \includegraphics[width=1.0\linewidth]{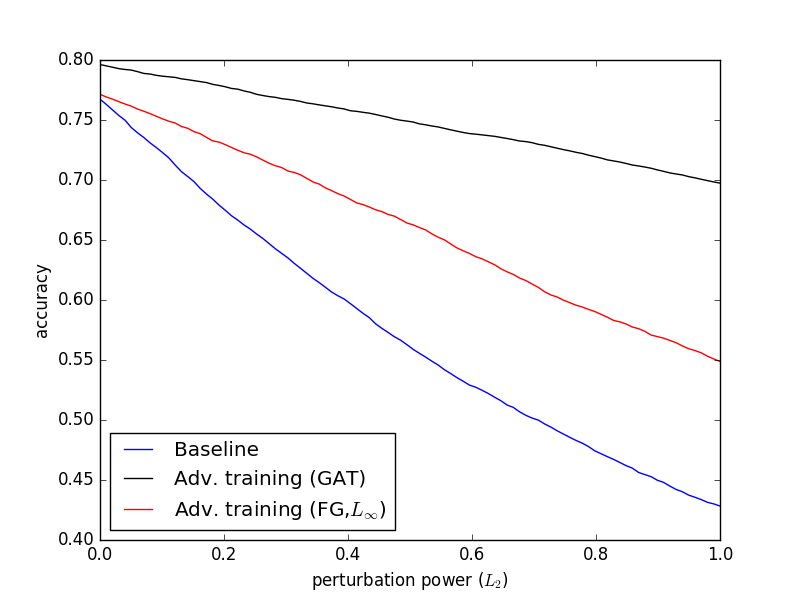}
  \caption{CIFAR-10}
\end{subfigure}%
\begin{subfigure}{.5\textwidth}
  \centering
  \includegraphics[width=1.0\linewidth]{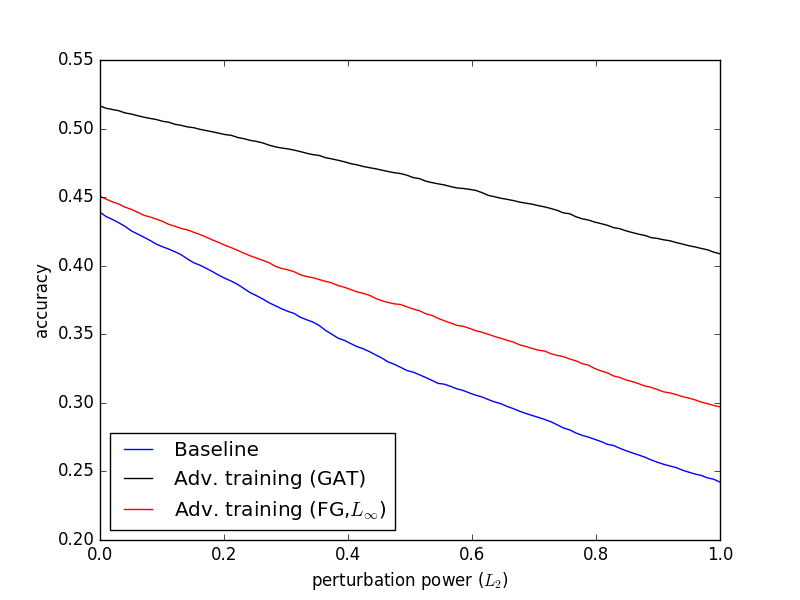}
  \caption{CIFAR-100}
\end{subfigure}
\caption{Comparison of the baseline and robustified networks. Adversarial examples were generated using (\ref{fgl2}) from another baseline network for various values of $\epsilon$, and classification accuracy is plotted as shown.}
\label{fig4}
\end{figure}

We next examined the robustness of the network against indirect attacks. Similar to previous experiments, an adversarial example was generated based on the fast gradient method with $L_2$ norm constraint. However, because attackers would be unfamiliar with the internal structure of each network, adversarial examples were generated using another baseline network. This kind of attack is effective because a large fraction of adversarial examples are misclassified by networks trained from scratch with different hyperparameters \citep{szegedy2013intriguing}. We compared the proposed adversarial training method with the baseline network and with the fast gradient method with $L_\infty$ norm constraint. The classification accuracy for the adversarial examples with various perturbation powers is shown in Fig. \ref{fig4}. It shows similar tendency as in the previous experiment, but classification accuracy did not decrease more rapidly in an indirect attack than in a direct attack.

\subsection{Regularization effect}
\label{regeff}

\citet{szegedy2013intriguing} showed that a neural network could be regularized through adversarial training. We performed an experiment to compare our method with dropout, random perturbation, and adversarial training using the fast gradient method. We used the set of hyperparameters that achieved the best performance on the validation data for each regularization method. We repeated this procedure 50 times using different weight initialization and obtained the average and standard deviation of the test accuracy. We measured the regularization performance on a low capacity network, which is intentionally used.  It is well known that there is more room for improvement in a low capacity network than in a high capacity network in terms of generalization error. It is also true that it is harder to improve the generalization error of a low capacity network than that of a high capacity network \citep[chap. 5]{Goodfellow-et-al-2016}. The results for the CIFAR-10 and CIFAR-100 datasets are shown in Table \ref{table2}.

The results show that the regularization effect of our method is remarkably superior to other methods. Existing regularization techniques show approximately $1 - 2\%$ performance improvement over the existing baseline network, whereas our method shows a remarkable $4 - 6\%$ improvement in accuracy. GAT can be applied to any neural network because it does not depend on the internal structure of a neural network. It appears that GAT is a powerful regularization technique that can be applied to any neural network.


\begin{table}[h]
\centering
\caption{Test accuracy for CIFAR datasets}
\begin{subtable}[b]{0.48\textwidth}
\caption{CIFAR-10}
\begin{tabular}{c c}
\hline
Method & Test accuracy(\%) \\
\hline
Baseline & 77.48 $\pm$ 0.46 \\
Dropout & 78.49 $\pm$ 0.64 \\
Random Perturbation & 77.59 $\pm$ 0.57 \\
Adv. training (FG, $L_\infty$) & 78.12 $\pm$ 0.59 \\
Adv. training (FG, $L_2$) & 77.99 $\pm$ 0.45 \\ 
\textbf{Adv. training (GAT)} & \textbf{80.33} $\pm$ \textbf{0.44} \\
\textbf{Dropout + GAT} & \textbf{81.62} $\pm$ \textbf{0.34} \\ 
\hline
\end{tabular}
\end{subtable}%
\hspace*{\fill}%
\begin{subtable}[b]{0.48\textwidth}
\caption{CIFAR-100}
\begin{tabular}{c c}
\hline
Method & Test accuracy(\%) \\
\hline
Baseline & 44.32 $\pm$ 0.63 \\
Dropout & 46.29 $\pm$ 0.61 \\
Random Perturbation & 44.43 $\pm$ 0.71 \\
Adv. training (FG, $L_\infty$) & 45.16 $\pm$ 0.73 \\
Adv. training (FG, $L_2$) & 45.67 $\pm$ 0.63 \\ 
\textbf{Adv. training (GAT)} & \textbf{50.44} $\pm$ \textbf{0.56} \\
\textbf{Dropout + GAT} & \textbf{50.71} $\pm$ \textbf{0.49} \\ 
\hline
\end{tabular}
\end{subtable}
\label{table2}
\end{table}

\section{Discussion}
We proposed a novel adversarial training method by combining adversarial training \citep{goodfellow2014explaining} and GAN \citep{goodfellow2014generative}. Experimental results show that our proposed method is not only robust against adversarial examples, but also effective in improving the generalization accuracy of the classifier. We believe that there are two main reasons to explain the better performance than the conventional fast gradient method.

First, the classifier has different robustness for each training data. In some images, the classifier can be easily fooled when having only low perturbation power. However, in other images, it cannot be easily fooled even with very high perturbation power. Because the conventional fast gradient method normalizes the size of each gradient, it generates adversarial images of the same perturbation power for all training images, which means that it is difficult for networks to converge. Generating adaptive adversarial examples based on the degree of robustness of each image can help efficiently train the network.

Second, the classifier network is a non-linear function. If the classifier network is a perfect linear function, finding better adversarial images than those found by the fast gradient method is impossible. However, because the classifier network is a non-linear function, GAT can detect non-linear patterns in the classifier network and use a gradient to produce better perturbations than with the fast gradient method.

Our proposed GAT effectively solves both problems because it does not normalize the gradient vector and evolves adaptively as the classifier is trained. Therefore, training a classifier that is robust to various adversarial examples is possible, and accordingly, effectively regularizes the model. However, further study is required because the proposed method takes $3$ to $4$ times longer training (depending on the capacity of the generator network) than the conventional fast gradient method. In addition, hyperparameters of each network should be carefully tuned because of the properties of GAN training.

\section{Conclusions}
In this paper, we proposed a method to make a classifier robust to adversarial examples using a GAN framework. The generator network generates a perturbation by finding the weaknesses of the classifier, and the classifier re-learns the image generated by the generator back to the original label. As the two networks learn alternately, the classifier network becomes more robust to the adversary image, and eventually the generator network will not be able to find a proper image that could fool the classifier. Our adversarial training method is also practical since it does not need expensive optimization process in the inner loop to find optimal adversarial images. The classifier with our adversarial training method is highly robust to the adversarial examples. Furthermore, it was found that the proposed method was surprisingly effective in regularizing neural networks.

To the best of our knowledge, this is the first method to apply a GAN framework to adversarial training (or supervised learning). Therefore, much work remains to improve the method. What is the optimal capacity of a generator network? Does additional information other than a gradient exist that can help the generator to find a better adversarial image? When our method is applied to larger networks such as Inception \citep{szegedy2016inception}, can similar results to those in this study be achieved? Further research is required to address these issues.

\bibliography{ref}

\end{document}